# Will I Get Hate Speech? Predicting the Volume of Abusive Replies before Posting in Social Media

Raneem Alharthi[a,*,1], Rajwa Alharthi[b], Ravi Shekhar[c], Aiqi Jiang[d] and Arkaitz Zubiaga[a]

[a]*Queen Mary University of London, United Kingdom*
[b]*Taif University, Saudi Arabia*
[c]*University of Essex, United Kingdom*
[d]*Heriot-Watt University, United Kingdom*

**ABSTRACT**

Despite the growing body of research tackling offensive language in social media, this research is predominantly reactive, determining if content already posted in social media is abusive. There is a gap in predictive approaches, which we address in our study by enabling to predict the volume of abusive replies a tweet will receive after being posted. We formulate the problem from the perspective of a social media user asking: "if I post a certain message on social media, is it possible to predict the volume of abusive replies it might receive?" We look at four types of features, namely text, text metadata, tweet metadata, and account features, which also help us understand the extent to which the user or the content helps predict the number of abusive replies. This, in turn, helps us develop a model to support social media users in finding the best way to post content. One of our objectives is also to determine the extent to which the volume of abusive replies that a tweet will get are motivated by the content of the tweet or by the identity of the user posting it. Our study finds that one can build a model that performs competitively by developing a comprehensive set of features derived from the content of the message that is going to be posted. In addition, our study suggests that features derived from the user's identity do not impact model performance, hence suggesting that it is especially the content of a post that triggers abusive replies rather than who the user is.

## 1. Introduction

The rapid proliferation of social media platforms, leading to a growth in the number of users globally, has made them an ideal source for online communication and interaction (Kwak, Lee, Park and Moon, 2010; Hughes, Rowe, Batey and Lee, 2012; Lozano-Blasco, Mira-Aladrén and Gil-Lamata, 2023). However, this is not without caveats, not least the use of such platforms for disseminating hateful speech (Walther, 2022). In part due to the anonymity offered by these platforms, users can easily post abusive messages attacking others, which leads to a distressing and unwelcome environment for many users. It fosters antagonism among users, which can escalate into serious real-world arguments and negatively impact individuals (Lupu, Sear, Velásquez, Leahy, Restrepo, Goldberg and Johnson, 2023). The severity of the issue has led to more research in developing automated methods for tackling hate online (Fortuna and Nunes, 2018; Yin and Zubiaga, 2021; Balayn, Yang, Szlavik and Bozzon, 2021).

Researchers have made significant progress in the automation of online hate speech detection methods by employing natural language processing (NLP) and text mining techniques. Recent studies, such as Schmidt and Wiegand (2017), highlight the urgent need for effective automatic detection systems. They showed the importance of combining diverse features, including surface-level, linguistic, and knowledge-based elements, while also emphasizing the potential of meta-information and multimodal data to enhance the accuracy and context-awareness of automated detection systems. Complementing this, Fortuna and Nunes (2018) emphasize the importance of combining various feature extraction methods to enhance detection accuracy and reliability. Other innovative approaches such as XP-CB (Yi and Zubiaga, 2022) and AnnoBERT (Yin, Agarwal, Jiang, Zubiaga and Sastry, 2023) further advance the field by addressing cross-platform generalization and inter-annotator agreement challenges, respectively. Yin and Zubiaga (2021) identify key obstacles such as non-standard grammar and dataset biases, proposing novel directions for future research to develop more robust hate speech detection systems, whereas others have looked at the generalizability

r.alharthi@qmul.ac.uk (R. Alharthi); raharthi@tu.edu.sa (R. Alharthi); r.shekhar@essex.ac.uk (R. Shekhar); a.jiang@qmul.ac.uk (A. Jiang); a.zubiaga@qmul.ac.uk (A. Zubiaga)
ORCID(s): 0000-0002-8798-641X (R. Shekhar); 0000-0002-9269-733X (A. Jiang); 0000-0003-4583-3623 (A. Zubiaga)

across languages, as is the case of Nandi, Sarkar, Mallick and De (2024) looking at the challenges of Indian languages, where code-mixing is common.

Researchers (Sorokowski, Kowal, Zdybek and Oleszkiewicz, 2020) have also incorporated psychological characteristics of users as a feature to detect online hate, finding that there are high levels of psychopathy, which can help distinguish individuals involved in hate speech from those engaging in other negative online behaviors such as trolling or cyberbullying. Álvarez-Carmona, Guzmán-Falcón, Montes-y Gómez, Escalante, Villasenor-Pineda, Reyes-Meza and Rico-Sulayes (2018) also incorporate author profiling, occupations and locations as features to enhance the aggressiveness detection, results shows an average F-measure of 0.6711. The role of anonymity and geographic variations were also incorporated as important features in Mondal, Silva and Benevenuto (2017) to enhance the detection and prevention of online hate speech. Additionally, Mishna, Cook, Gadalla, Daciuk and Solomon (2010) used features like the prevalence of bullying among youth, relationships between perpetrators and victims, emotional responses, and motivations for bullying. These features enhanced the prediction of cyberbullying and provided a nuanced understanding of the problem.

However, as recent research continues to highlight the challenges in detecting hateful content that has already been posted (Rawat, Kumar and Samant, 2024), there is a gap in research when it comes to proactively anticipating if a message, once posted, will be replied to by others with hate speech, i.e. predicting the volume of abusive replies a post will receive before posting it, so that one can be prepared in advance. We envisage a user asking: "If I post this message as is on social media, will it be responded to with abusive replies?" We aim to assist the user in being prepared before posting a message, enabling support advising how to do it best to minimize the volume of abusive replies received, as well as in furthering understanding into mitigating abusive content received.

Currently, most efforts focus on classifying already posted content as hate or non-hate. The ability to anticipate and consequently prevent potential hateful responses could provide valuable insights and enable users to make informed decisions about how to post in social media knowing the potential of a message to receive abusive replies. We experiment with the ability to achieve this through comprehensive experiments testing the effectiveness of different features in determining the volume of abuse a post will receive. Given that we only rely on features available at the time of posting and hence exclude other features such as the number of interactions (e.g., retweets, likes) which develop over time after posting and hence are only available at later stages after the message is already posted.

Our main objective is to test the predictions made by our designed feature sets to predict how much abuse a post will receive as a regression task. To address this objective, we define and tackle the following research questions (RQs):

- **RQ1:** What categories of features available at the time of posting are helpful in predicting the volume of abusive replies a tweet will receive?

- **RQ2:** To what extent is the volume of abusive replies linked to the identity of the user posting a message or the content of the message being posted?

We address these RQs through the following regression experiments. We do this by using a range of machine learning algorithms in the interest of the generalizability of our analysis. Using the Online Abusive Attacks (OAA) dataset (Alharthi, Alharthi, Shekhar and Zubiaga, 2023), we experiment with a broad set of features to investigate the problem of predicting the potential for abusive replies and to answer the RQs above towards a better understanding of what enables a better prediction. Our study provides, in turn, a better understanding of what triggers abusive replies, suggesting possible ways of improving the design of social media platforms to reduce the volume of abuse.

*Contributions.* The main contributions of our study are as follows:

- We perform regression experiments aiming to predict the number of abusive replies a social media post will receive. To answer RQ1, we formulated it as a regression problem to predict the number of abusive replies to a social media post. Our experiments show that we can predict the volume of abusive replies with good accuracy.

- Our study performs an innovative study looking at the effectiveness of different features in the predictive experiments, helping understand what triggers abuse in social media, as well as advancing research towards helping users determine if a message they plan to post is likely to receive a certain amount of abusive replies before they go ahead and post it.

*Findings.* Our study finds that one can achieve highly competitive performance in predicting the volume of abuse a message is likely to receive, which we measure as the number of abusive replies. Our results consistently show that, where one wants to predict if a post will receive abusive replies, it is much more reliable to extract features from the content of the post, while features derived from the posting user's identity do not have an impact on the performance; this indicates that it is especially the content of a post which is predictive of abusive replies rather than who the user is.

*Paper structure.* This article is organized as follows. The following section reviews related work, including topics such as predicting the spread and intensity of abusive language online, understanding the role of perpetrators in abusive language detection, and mitigating abusive language. Then, we delve into our methodology, describing the dataset used, the models developed, and the steps taken for text preprocessing, feature extraction/engineering, and experiment settings. We also detail the problem formulation, training details, and evaluation metrics used. Our research proposes a regression task to predict the volume of abuse a tweet will eventually receive. The paper concludes with a discussion of our findings and a final conclusion.

## 2. Related Work

Automated detection of online hate is a growing research area in recent years (Jiang and Zubiaga, 2024; Gandhi, Ahir, Adhvaryu, Shah, Lohiya, Cambria, Poria and Hussain, 2024), due to the pressing need to tackle and mitigate online abuse (Saha, Ahlawat, Akram, Jangbahadur, Dhaigude, Sharma and Kumar, 2024; Davidson, Grove-Hills, Bifulco, Gottschalk, Caretti, Pham and Webster, 2011; Watson, 2024). In this section, we discuss existing research whose focus has been on performing different kinds of classification task to support the process of online hate detection, followed by the recent research that focuses on the perpetrators and how that contributed in the field, and finally attempts to mitigate online abuse by incorporating features that provide context awareness to the online hate event in which results in deep understanding of the problem. We then discuss the need for our proposed research to predict the potential for abuse before posting on social media.

### 2.1. Analyzing and predicting the spread and intensity of abusive language online

There has been a body of research that has been studying and attempting to investigate how abusive content spreads (Gashroo and Mehrotra, 2024), as well as determining its intensity or severity (Goel and Poswal, 2024).

Researchers in Mathew, Dutt, Goyal and Mukherjee (2019) investigate how hate speech spreads on Gab. By analyzing 21 million posts from 341,000 users, the researchers found that hateful content spreads faster and reaches a wider audience than non-hateful content. Despite representing only 0.67% of the users, hateful users were responsible for 26.80% of the posts, facilitated by their dense network connections. The study employs a lexicon-based approach and the DeGroot model, which was used to identify users who are potentially spreading hate speech, using a belief repost network and running a DeGroot model diffusion process. While the focus on Gab limits generalizability, the research provides critical insights into the mechanisms of hate speech spread online.

Masud, Dutta, Makkar, Jain, Goyal, Das and Chakraborty (2021) explore the generation and spread of hate speech on Twitter. It introduces RETINA, a neural architecture that uses exogenous signals to predict retweet dynamics, achieving a macro F1-score of 0.85. The study highlights that hate speech is often topic-dependent and more retweeted than non-hateful content, though it creates fewer susceptible users over time. While the study provides significant insights and a robust model, it primarily focuses on retweet cascades, which may not fully capture the complexity of hate speech diffusion.

The authors in Dahiya, Sharma, Sahnan, Goel, Chouzenoux, Elvira, Majumdar, Bandhakavi and Chakraborty (2021) address the challenge of predicting the evolution of hate speech in Twitter reply threads. By proposing the DESSERT model, which merges neural network capabilities with statistical signal processing, the study effectively forecasts hate intensity, outperforming existing models with three layers achieved a Pearson correlation coefficient of 0.670, outperforming the best baseline model, ForGAN, which had a Pearson correlation coefficient of 0.557. The research highlights that hate intensity in replies is not necessarily linked to the source tweet's tone, emphasizing the need for predictive models in content moderation. Despite its strengths, the model's dependency on specific datasets and classifiers suggests areas for further exploration and validation.

Sahnan, Dahiya, Goel, Bandhakavi and Chakraborty (2021) address the challenge of determining the hate intensity in Twitter reply chains using a novel model called DRAGNET. By employing a deep stratified learning approach,

DRAGNET categorizes heterogeneous hate profiles into homogeneous clusters, significantly outperforming existing baseline models including ForGAN with a 9.4% higher Pearson correlation coefficient and a 19% lower Root Mean Square Error (RMSE).

Mondal et al. (2017) present a dataset and a system for estimating abuse severity and target in online social media posts. They focus on the presence, severity, and target of abusive behavior, achieving high accuracy for abuse presence, abuse target prediction, and abuse severity classification.

Chandra, Pathak, Dutta, Jain, Gupta, Shrivastava and Kumaraguru (2020) introduce a pioneering dataset of 7,601 Gab posts, labeled for presence of abuse, severity, and target. By employing both traditional machine learning and deep learning models, the study achieves high accuracy in predicting these dimensions, with BERT emerging as the most effective model with 80% accuracy for abuse presence, 82% for target prediction, and 65% for severity classification. While the study provides valuable insights into the nature of online abuse, particularly on Gab, it faces challenges like class imbalance and the nuanced nature of some abusive content.

The advancement in research methodologies, including diffusion analysis and intensity classification models, offers innovative approaches to anticipate and manage online hate speech more effectively. These solutions not only enhance the ability to identify and address instances of hate speech but also empower us to include more predictive features to enhance online mitigation (Chakraborty and Masud, 2022; Saha, Das, Mathew and Mukherjee, 2023; Chung, Abercrombie, Enock, Bright and Rieser, 2023; Kumar and Maurya, 2024). Still, all this research focuses on determining the intensity of abuse, but has not looked at predicting the volume of abuse before posting in social media.

## 2.2. The role of perpetrators in abusive language detection

The following studies tackled the online hate detection problem by studying and analyzing the hate users "perpetrators", behavior and characteristics.

Ribeiro, Calais, Santos, Almeida and Meira Jr (2017) propose a user-centric view of hate speech to enhance detection methods and understanding. The research collected a Twitter dataset of 100,386 users, with 4,972 users manually annotated as hateful or not through crowdsourcing. They found that hateful users exhibit distinct activity patterns compared to normal users, including more recent account creation dates, increased statuses and followers per day, more favorites, shorter tweet intervals, and greater centrality in the retweet network. Additionally, hateful users are characterized by more negative and profane language use. They also find that these users are more central in the retweet network compared to normal users. This is a counterintuitive finding because it contradicts the "lone wolf" stereotype, which suggests that individuals engaging in hateful behavior are isolated or less connected.

Sorokowski et al. (2020) aimed to compare individuals who post hate comments online with those who do not. The research involved 94 Internet users, with findings indicating that high scores in the Psychopathy subscale were significant predictors of posting hating comments online, while high scores on the Envy Scale were marginally significant. This study provides initial evidence that individuals engaging in derogatory online behavior exhibit a high level of Psychopathy with a logistic regression coefficient of 1.37 and a statistically significant result ($p < 0.001$). but do not necessarily have elevated levels of other traits commonly associated with disruptive behavior. It distinguishes online hating from other derogatory online behaviors like trolling, cyberbullying, or hate speech, shedding light on the psychological background of online haters.

ElSherief, Nilizadeh, Nguyen, Vigna and Belding (2018) provide a comprehensive analysis of hate speech dynamics on Twitter. By curating a large dataset with 27,330 hate speech tweets, identifying 25,278 instigator and 22,857 target accounts, the authors identify key differences between instigators and targets in terms of visibility and personality traits. The study reveals that hate speech participants are more active and visible on Twitter and share distinct personality characteristics and have more followers, retweets, and lists than general Twitter users. These findings contribute to the understanding of online hate speech and offer valuable resources and methodologies for future research.

Analyzing the entire hate incident is considered an effective approach that offers a comprehensive overview of the contributing key factors. Thus, researchers started to propose context-aware online hate speech detection models.

Mosca, Wich and Groh (2021) investigate the impact of user context on hate speech detection models, revealing that user features can enhance model performance.

By delving into these studies, we gain insights into the multifaceted nature of online hate speech, exploring how different features play a crucial role in shaping the dynamics of online hate incidents. Understanding the context in which hate speech occurs and the specific contributions of each individual involved can provide valuable information for devising targeted interventions and preventive measures.

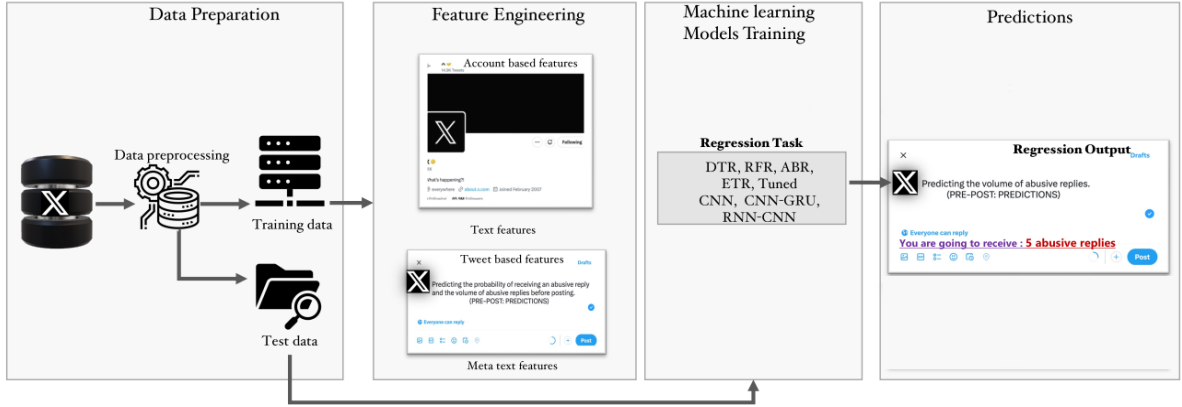

Figure 1: An overview of the proposed framework for the regression tasks.

### 2.3. Mitigating abusive language by incorporating context aware features

A novel task of hate speech normalization is addressed in Masud, Bedi, Khan, Akhtar and Chakraborty (2022), which aims to reduce the severity of hate speech while providing users with a stepping stone towards non-hate behavior. They manually curate a parallel corpus of hate texts and their normalized counterparts, proposing the NACL model. This model measures hate intensity, identifies hate spans, and reduces hate intensity by paraphrasing hate spans. NACL outperformed six baselines, achieving a root mean square error (RMSE) of 0.1365 for intensity prediction, an F1-score of 0.622 for span identification, and BLEU and perplexity scores of 82.27 and 80.05, respectively, for text normalization.

Valle-Cano, Quijano-Sánchez, Liberatore and Gómez (2023) present SocialHaterBERT, a novel model for detecting hate speech on Twitter by combining textual analysis with user profile features. Building on the BERT-based HaterBERT model, it incorporates a Social Graph methodology to analyse user interactions and characteristics. This integrated approach significantly improves detection accuracy, achieving a 4% increase in F1-score over text-only models. While the model shows promising results, its reliance on user data and focus on Spanish datasets are noted limitations.

As shown above researchers have focused on measuring the diffusion and intensity of abuse, considering author characteristics in the detection process, categorizing hateful users, and identifying the roles of individuals involved. Despite these efforts towards furthering online hate detection, there is a clear gap in ability to predict the volume of the abusive replies a post might receive before being posted. To the best of our knowledge, there is no research investigating the ability to predict whether, and the extent to which, a future post is likely to receive abuse; this is however crucial as an effort to protect social media users, by flagging the risk in advance, and enabling to decide how to post content in social media.

### 3. Methodology

In this section, we describe the methodology of our research, including the resources we use and the approach we define to tackle the abuse prediction task, which is depicted in Figure 1.

### 3.1. Problem Formulation

Let $\mathcal{T} = \{\tau_1, \ldots, \tau_i, \ldots, \tau_N\}$ represent a collection of tweets, where each tweet $\tau_i$ is a post containing a text and its associated metadata, posted by a user and which is part of a conversation, i.e. is replied to by other replying tweets $\mathcal{R}_i = \{r_{i1}, \ldots, r_{in}\}$. Each replying tweet in $\mathcal{R}$ is labelled as abusive or not abusive, and from these labels we can derive the volume of abuse received by $\tau_i$ as the number of replying tweets that are labelled as abusive, where the volume of abuse is defined as the target label $y_i$.

The predictive regression function $f(\chi_i)$ is defined to predict the volume of abuse that a post will get, i.e. $\hat{y}_i$, where the aim is to minimize the error of the prediction by reducing the error between $\hat{y}_i$ and $y_i$. We use a supervised learning approach to develop a regression method that can approximate $y_i$ to predict the volume of abusive replies a tweet will get.

Features such as tweet text content and tweet metadata are crucial for training models. During training, different combinations of these features are used as inputs for the models used for the experiments to effectively capture the correlation between the predictive features and the volume of abusive replies.

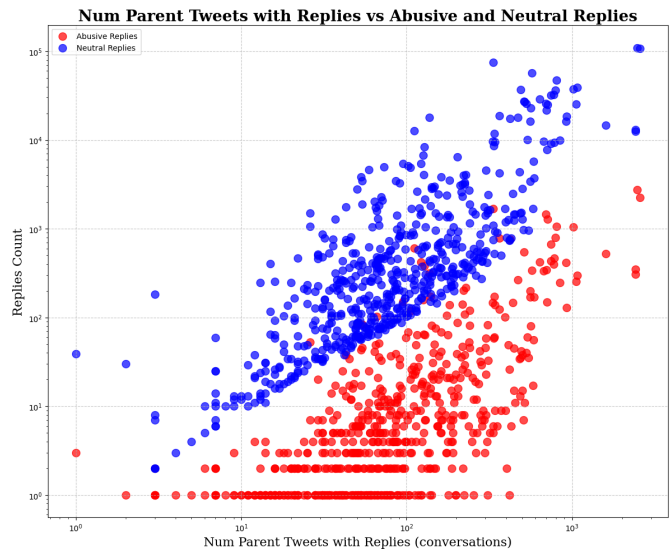

Figure 2: Relationship between conversation frequency (x-axis) and reply count (y-axis) using a logarithmic scale. Red dots represent abusive replies, while blue dots indicate neutral replies. The logarithmic y-axis highlights relative differences in reply volume across conversations.

### 3.2. The Online Abusive Attacks (OAA) Dataset

We next describe the OAA dataset we use in our study, the data sampling strategy focus on our objectives, and the resulting distribution of the dataset.

***The OAA dataset.*** We use our previously published dataset, the large-scale and diverse Online Abusive Attacks (OAA) dataset[1] (Alharthi et al., 2023), which comprises 2.3K Twitter accounts, 5 million tweets, and 106.9K categorized conversations. A conversation refers to a parent tweet that has at least one reply from another user.[2] The dataset contains 153,144 first level replies. With an average of 45.1 conversational tweets per target and a weighted average of 14.3 replies across all conversations of different targets.

Focusing on the targets of abuse, rather than just the abusive content itself the target-centric data collection involved four phases: topic selection, target identification, target-centric data collection, and metadata collection. Initially, popular and contemporary topics were identified using trending hashtags, followed by the selection of potential targets of abuse based on their activity on these topics. The dataset was further enriched by collecting comprehensive metadata for the accounts and tweets involved, ensuring a diverse and inclusive dataset beyond predefined keywords or hashtags Alharthi et al. (2023).

The annotation process involved labelling tweets using Google Jigsaw's Perspective API, which assigns toxicity scores based on attributes like identity attack, insult, and profanity. Tweets exceeding a certain threshold were marked as abusive. Manual validation by annotators confirmed the reliability of this automated labelling.

Figure 2 shows the distribution of the abusive and neutral replies across conversation volumes of targets where the red points represent the abusive replies and blue points represent neutral replies. It clearly shows that the neutral replies are increasing more rapidly with the number of conversations.

***Sampling tweets with at least one reply.*** The original OAA dataset contains a large number of parent tweets that received no replies. For the purposes of investigating whether the replies attracted by tweets are abusive, in this work, we use the subset of the dataset made by parent tweets with at least one reply, which are 106,914 parent tweets used for the final dataset. We are also considering only direct replies, which are responses made to a specific tweet that is directed at the original sender; hence, deeper replies in the conversations, including second-level replies, are ignored. Both labelled parent tweets and labelled replies are utilized. The collected information includes a comprehensive user data of social media profiles for the target as a longitudinal data, and 144 distinct metadata for both parent tweet and the direct replies, which provides contextual information about the abusive attacks enabling in-depth analysis of factors contributing to abuse.

---

[1] https://github.com/RaneemAlharthi/Online-Abusive-Attacks-OAA-Dataset
[2] https://help.x.com/en/using-x/x-conversations

*Using the OAA dataset in our study.* The overall aim is to estimate the volume of abusive replies, which we achieve through a regression model. The experiments will be individually described in detail in the following sections. Table 1 provides the final statistics of the OAA dataset as used in the present study, where we can see that 20% of the parent tweets receive at least one abusive reply, whereas the remainder 80% receives replies but where none is abusive.

Table 1: Statistics of the final OAA dataset as used in our study.

| Feature | Count |
| --- | --- |
| Number of user accounts | 2,367 |
| Number of conversations | 106,914 |
| Number of conversations with abusive replies | 21,383 (20%) |
| Number of conversations with non-abusive replies | 85,531 (80%) |
| Number of replies | 153,144 |
| Number of abusive replies | 24,907 (16.3%) |
| Number of non-abusive replies | 128,237 (83.7%) |

The dataset contains a holistic collection of conversations incorporating user and textual features, which we group into four types of features for our experiments, which we describe later.

### 3.2.1. Text Preprocessing

Prior to training the models, we conduct the following preprocessing to clean and transform all the features that are available in textual format, which includes the text of the tweets and user descriptions, including the biography of users. We use the following functions from the scikit-learn, spaCy and NLTK python packages: 'OneHotEncoder', 'StandardScaler', and 'Tokenizer', to perform the following preprocessing steps:

1. **Tokenization:** We tokenize the texts into individual space-separated tokens.
2. **Stopword Removal:** We use pre-defined libraries such as NLTK[3] (Natural Language Toolkit) and spaCy.[4]. We also include our own set of stop-words, identified during a manual inspection from the training data, especially during the generation of word clouds, such as 'is', 'at', 'the', 're', 'name', 'user', 'ct', 'us', 'ud', 'ua', 'ut', 'amp', 'uc', 'ue', 'uk', 'it', 'im', 'ut'.
3. **Stemming:** We use a stemmer to reduce words to their base or root forms, enhancing the efficiency of the classification model.
4. **Special Character Removal:** including:
   - Punctuation marks, symbols, and others that are not a word character or a whitespace character,</>//@/etc.
   - Non-ASCII characters (including emojis, certain special characters, accented letters, and other symbols outside the standard ASCII range).
   - Extra spaces (including multiple consecutive spaces and leading and trailing spaces).
   - Unicode numbers.
   - Single-letter words.

---

[3] https://www.nltk.org/
[4] https://realpython.com/natural-language-processing-spacy-python/

*3.2.2. Feature Extraction/engineering*

After preprocessing, we extract text-related features using Bag of Words (BOW). We limited the vector size of the vocabulary to the top 5,000 most frequent words, effectively reducing the dimensionality of the feature space. Additionally, we incorporate various lexicons in our analysis. As they act as dictionaries of words linked to specific categories, these lexicon words are used to construct binary features by counting their occurrences in text data, with each word representing its presence or absence. The process involves iterating through each text in a dataset, counting lexicon word occurrences, and calculating a score.

The BoW vectorization captures the word occurrences in the sentence, while the lexicon-based features add sentiment information to the feature set. By combining these two types of features, the models can consider both the presence of words and their sentiment polarity to improve prediction accuracy.

We define four types of different features, given our interest in exploring how each of them affects the prediction performance. Hence, we perform experiments in different settings incorporating or excluding different subsets of these four feature types. The four feature types include:

1. **Text features (Te)**: The text features include all text presented in the captured context of a complete sample, which includes the text in the parent tweet, replies, and the user's bio description using the following lexicons.
   (a) Lexicon of Abusive Words:[5] The "Lexicon of Abusive Words" by UDS-LSV is a GitHub repository containing a comprehensive collection of abusive words, phrases, and expressions from various languages. It is structured to include different languages, each with its own set of abusive words and phrases, categorized by the type of abusive language (e.g., hate speech, insults, threats).
   (b) Hatebase Dictionary:[6] [7] We used the provided tool that enables the extraction of a comprehensive list of English hate speech terms for analysis and detection from the Hatebase Dictionary.
   (c) Positive Words/Opinion Sentiments:[8] The English opinion lexicon, as well as lists of negative and positive words were used.
   (d) Negative Words Lexicon:[9] We used the negative and abusive list.
   (e) GloVe lexicon, Global Vectors for Word Representation, which is employed to create word embeddings for the DL models.[10]

2. **Text meta features (Mt)**: It includes all additional information and attributes associated with the text without providing the exact text, such as stemmed character, hate word counts, negative word counts, positive word counts, abusive word counts, character count of parent tweet.

3. **Tweet based features (Tw)**: Tweet based features are the features related to the tweet and the text of the tweet both in the (pre-post) and (after post) stages: the pre-post tweet refers to a tweet in a draft state: The tweet exists in a draft form, composed but not yet publicly visible. and the after post tweet is the published tweet, a publicly visible tweet. such as hashtags, mentions, hate, abuse in the text content, etc.

4. **Account based features (Ac)**: Account based features are the features that describe the user's account, such as follower count, favorite tweet count, etc. This group of features enables us to assess to which it is the user's characteristics that motivate others to post abusive replies to them, or it is instead the posts, as captured by the other three feature sets.

For the textual features, both Bag of Words (BoW) and TF-IDF were initially utilized in the machine learning models, but only BoW was chosen for further analysis after we observed that both approaches led to similar results. We also exclude the textual description from the users' bios, as initial empirical results suggested that this feature did not make a difference in the overall performance of the models, both when utilized independently and when combined with other features. In the interest of focus and brevity in the paper, we exclude these results from further analysis.

We provide a detailed list of features for each category, in the appendix A. The feature sets mentioned above serve the primary purpose of being used for prediction and offer valuable insights into the underlying reasons behind abusive attacks.

---

[5] https://github.com/uds-lsv/lexicon-of-abusive-words
[6] https://hatebase.org/
[7] https://gist.github.com/noahgh221/68ae96c3f6dc41c9596fed704fdc454d
[8] https://github.com/jeffreybreen/twitter-sentiment-analysis-tutorial-201107
[9] https://github.com/usc-sail/mica-violence-ratings
[10] https://nlp.stanford.edu/projects/glove/

---

## 3.3. Regression Models

This section presents our models, feature engineering process, text preprocessing, and NLP techniques.

We started by selecting suitable models to perform the regression experiments. The final goal of these experiments is to predict the volume of abusive replies (regression). Interpretability is crucial to understanding the relationship between the selected predictive features and the output prediction. Thus, more interpretable models, such as linear Regression Decision Trees and Random Forest were selected as the primary models based on their ability to provide clear insights into how each feature contributes to the prediction. Each of these models has its strengths and is chosen based on each task's specific requirements, the dataset's size, the importance of capturing context, and the need for interpretability versus performance.

For the regression experiments, we use the following algorithms:

- **RandomForestRegressor:** Random forests are chosen for their ability to handle high-dimensional data and missing values. They are robust and relatively easy to use, requiring minimal tuning of hyperparameters.

- **AdaBoostRegressor:** AdaBoost is selected for its effectiveness in improving the performance of weak learners by focusing more on the instances that are harder to classify. This adaptive boosting approach can lead to a model that is more robust and accurate.

- **ExtraTreesRegressor:** Extra Trees Regressor is chosen for its ability to handle a large number of features and its robustness against overfitting. It's an ensemble method that combines multiple decision trees, making it effective for capturing complex patterns in the data. This model is particularly useful for regression tasks where the data has a high dimensionality.

- **Tuned CNN Model Regressor:** CNN models are selected for their ability to capture local patterns in the data, which is crucial for understanding the context and semantics of the data. The tuned CNN model regressor is optimized for the specific characteristics of the regression task.

- **Tuned CNN-GRU Model Regressor:** The combination of CNNs for feature extraction and GRUs for capturing sequential information makes the CNN-GRU model regressor highly effective for tasks that require understanding the context of words in a sentence, which is crucial for regression tasks involving sequential data.

- **Tuned Attn Based Optimized RNN-CNN Model Regressor:** This model combines the strengths of RNNs for sequence modeling and CNNs for feature extraction with an attention mechanism to focus on the most relevant parts of the input.

- **BERT model:** The pre-trained transformer-based model, specifically 'bert-base-uncased' is selected for handling the regression tasks involving text data due to its bidirectional nature, which allows it to capture rich contextual information from both directions in the input text. We have modified the model's architecture in order to understand complex relationships between words and their context alongside with the different metadata categories in our work, making it effective for predicting both binary and continuous outcomes based on textual input and the extra metadata layer. In this work, the BERT model was fine-tuned on the OAA dataset, adapting its pre-trained language understanding to the nuances of the regression tasks. The model's output is combined with additional meta-features layer, allowing it to leverage both textual and numerical information

    Figure 3 depicts the BERT model architecture that combines BERT embeddings with additional meta-features for the regression task. The downstream process begins with the BERT layer, which processes the input text (input ids, attention mask, and token type ids) to output contextual embeddings. These embeddings are then mean-pooled to create a fixed-size representation. Separately, the meta-features are processed through their own input layer. The BERT embeddings and meta-features are concatenated, forming a rich representation of both textual and non-textual data. This combined representation then passes through a dense layer with ReLU activation, followed by a dropout layer for regularization. Finally, the output layer consists of a single neuron for regression prediction.

    Different from the other models, as the BERT model is dependent on the text input, we do not run the model with feature sets excluding text. Instead, we experiment with all possible alternatives that include text in addition to permutations of other features.

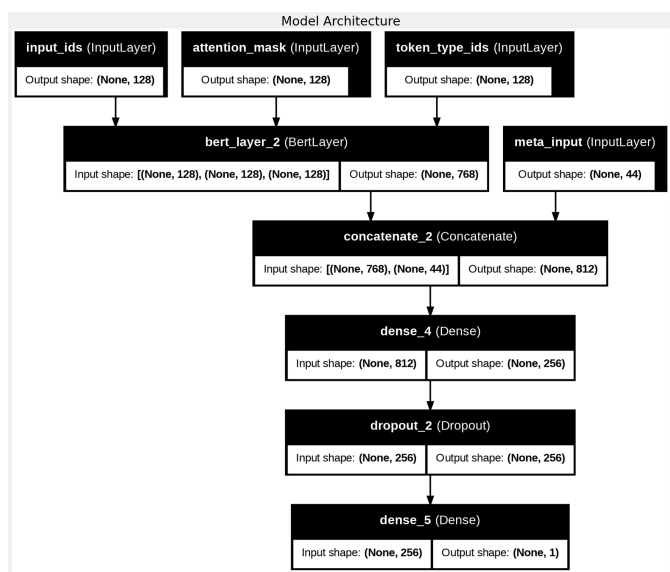

Figure 3: A visualisation of the BERT model architecture

## 4. Experiment Settings

The dataset exhibits a left-skewed distribution, characterized by a longer tail on the left side. This means that most data points are concentrated on the right, with fewer extreme values on the left. To effectively manage this variability across all features in our experiments, we applied scaling to the numerical features. Specifically, we utilized standardization as the scaling method, which ensures that each feature has a mean of zero and a standard deviation of one, thereby normalizing the data and mitigating the effects of skewness. A normalization process was applied to ensure comparability across texts of varying lengths, and then we introduced a threshold of 0.1 to classify texts as abusive or non-abusive. The threshold of 0.1 for classifying tweets as "hatred/abusive" is chosen based on several key factors. Empirical analysis showed that most tweets score low, making 0.1 an effective point to identify more extreme cases. This threshold balances the risks of false positives and false negatives, ensuring that tweets are accurately classified. It indicates that at least 10% of the words in a tweet are abusive, reflecting a clear intent to spread hate. This threshold allows us to have a balance between capturing abusive content and minimizing false negatives. If a score exceeds 0.1, it is classified as abusive. However, if the score is exactly 0.1, we flag it for manual inspection of cases where negative words might have skewed the score toward the abusive threshold.

### 4.1. Training details

As we are dealing with abusive replies detection where the classes are imbalanced (20% of conversations with at least one abusive reply, 80% of conversations without abusive replies), it is crucial to maintain the same class distribution in both training and testing sets to prevent bias in the model's performance evaluation.

Therefore, random sampling is used to split the data into training and testing sets. We used SMOTE (synthetic minority over sampling technique) to balance the classes, which generated more samples in the training set.

***Hyperparameter tuning*** A hyperparameter tuning was done for each model and was tuned using trial and testing to get the best possible scenario.

### 4.2. Evaluation Metrics

The mean of MSE and R-squared are used to evaluate the regression task models in a five-fold cross validation methodology.

***Test MSE (Mean Squared Error).*** Mean Squared Error measures the average squared difference between estimated values and actual values.

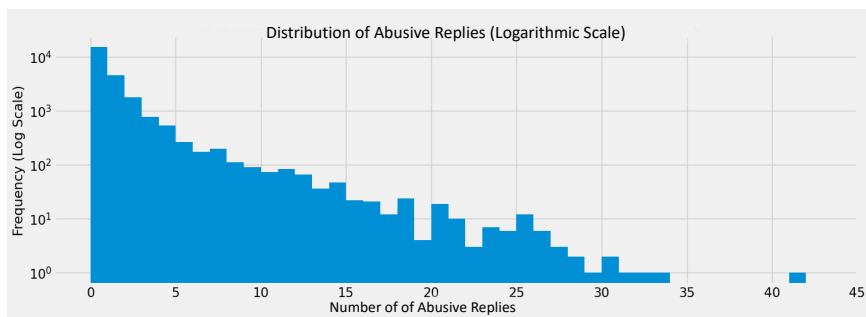

Figure 4: A distribution of number of replies, where the x-axis represents the number of abusive replies, and the y-axis represents the frequency of occurrence in a logarithmic scale.

***R-squared.*** R-squared ($R^2$) is a statistical measure that explains the proportion of variance in a dependent variable explained by the model. It is calculated using the sum of squared errors.

***Training procedure.*** For the deep learning models, we use the following settings: 30 epochs and no early stopping; early stopping and LR scheduler was used in the iterations only.

To enhance the interpretability of our machine learning model and gain insights into feature importance, we employed SHAP (SHapley Additive exPlanations) analysis, which helps identify the saliency of features across predictions.

## 5. Predicting the Volume of Abuse: Results

Next, we present and discuss the results of the regression experiments attempting to predict the number of abusive replies a social media post will get before it is posted, by looking at different regression models and different permutations of features. These results provide insights into the potential of predicting the volume of abuse as well as into the effectiveness of different features.

***Distribution of the volume of abusive replies.*** Before getting into the analysis of results, we analyze the frequency of different volumes of abusive replies. Figure 4 shows the number of replies and their frequency in a logarithmic scale. This suggests that the number of abusive replies is skewed towards lower values, with decreasing numbers of conversations receiving larger numbers of abusive replies. However, it can be seen that higher number of abusive replies are also frequent, and that some conversations can receive over 30 and over 40 abusive replies. This demonstrates the challenging nature of the task of predicting the volume of abusive replies given the wide distribution of values.

***Analysis of results.*** In Table 2, we present the results of the regression experiment that was performed to assess the ability to predict the volume of abusive replies a tweet might get using all the previously listed features. We measure the performance using the mean of R-squared (higher is better) and Mean Square Error (MSE, lower is better) metrics.

***Comparison of models.*** Looking at the seven regression models under study, we observe that the three models on the left (RFR, AR, ETR) perform better than the rest of the models in their best case. These three models are the only ones achieving MSE scores below 0.70, where MSE scores for the rest of the models are above 0.90 (CNN-RNN) and above 1.00 (CNN-FCN, CNN-GRU, BERT). Likewise, the same three models (RFR, AR, ETR) achieves the highest $R^2$ scores between 0.88 and 0.89, where other models perform in the range between 0.79 and 0.84 in the best case.

Table 2: Performance scores for the regression experiments predicting the volume of abuse. Best (or joint best) results for each model highlighted in bold, with the best overall also underlined. The BERT model includes only results for combinations of features including text, given its dependency on text which is then concatenated with other features.

| | Features | | | | RFR | | AR | | ETR | | CNN-FCN | | CNN-GRU | | CNN-RNN | | BERT | |
|---|---|---|---|---|---|---|---|---|---|---|---|---|---|---|---|---|---|---|
| # | Te | Mt | Tw | Ac | MSE | $R^2$ | MSE | $R^2$ | MSE | $R^2$ | MSE | $R^2$ | MSE | $R^2$ | MSE | $R^2$ | MSE | $R^2$ |
| 1 | X | | | | 1.52 | 0.72 | 1.60 | 0.70 | 1.49 | 0.76 | 3.79 | 0.37 | 3.82 | 0.36 | 3.89 | 0.35 | 4.90 | 0.18 |
| 2 | | X | | | 0.87 | 0.85 | 0.89 | 0.85 | 0.83 | 0.86 | 1.44 | 0.76 | 5.84 | 0.037 | 6.02 | 0.00 | – | – |
| 3 | | | X | | 1.00 | 0.83 | 1.05 | 0.82 | 0.97 | 0.83 | 1.91 | 0.68 | 2.42 | 0.02 | 5.87 | 0.03 | – | – |
| 4 | | | | X | 6.07 | -0.00 | 6.81 | -0.12 | 6.07 | -0.00 | 6.07 | -0.00 | 6.07 | -0.00 | 6.07 | -0.00 | – | – |
| 5 | X | X | | | 0.87 | 0.85 | 0.92 | 0.84 | 0.83 | 0.86 | 1.25 | 0.79 | 1.25 | 0.79 | 1.22 | 0.79 | 1.38 | 0.76 |
| 6 | X | | X | | 0.95 | 0.84 | 0.98 | 0.83 | 0.88 | 0.85 | 2.98 | 0.50 | 3.04 | 0.49 | 1.75 | 0.71 | 2.13 | 0.64 |
| 7 | X | | | X | 4.85 | 0.20 | 19.80 | 0.28 | 4.50 | 0.27 | 3.78 | 0.37 | 3.78 | 0.37 | 3.93 | 0.35 | 4.90 | 0.18 |
| 8 | | X | X | | 0.68 | 0.88 | 0.69 | 0.88 | **<u>0.62</u>** | **<u>0.89</u>** | 1.15 | 0.80 | 5.82 | 0.04 | 6.02 | 0.00 | – | – |
| 9 | | X | | X | 0.87 | 0.85 | 0.90 | 0.85 | 0.82 | 0.86 | 1.46 | 0.75 | 5.91 | 0.025 | 6.02 | 0.00 | – | – |
| 10 | | | X | X | 1.00 | 0.83 | 1.09 | 0.81 | 0.97 | 0.83 | 1.90 | 0.68 | 6.01 | 0.00 | 5.88 | 0.03 | – | – |
| 11 | X | X | X | | **0.67** | **0.88** | 0.70 | 0.88 | 0.64 | 0.89 | 1.22 | 0.79 | **1.21** | **0.79** | 1.09 | 0.81 | 1.21 | 0.79 |
| 12 | X | X | | X | 0.87 | 0.85 | 0.92 | 0.84 | 0.84 | 0.86 | 1.27 | 0.79 | 1.25 | 0.793 | 1.11 | 0.81 | 1.40 | 0.76 |
| 13 | X | | X | X | 0.95 | 0.84 | 1.04 | 0.82 | 0.89 | 0.85 | 3.06 | 0.49 | 3.01 | 0.50 | 1.98 | 0.67 | 2.12 | 0.64 |
| 14 | | X | X | X | **0.67** | **0.88** | 0.68 | 0.88 | 0.63 | 0.89 | **1.12** | **0.81** | 5.89 | 0.02 | 6.03 | 0.00 | – | – |
| 15 | X | X | X | X | **0.67** | **0.88** | **0.68** | **0.88** | 0.64 | 0.89 | 1.23 | 0.79 | 1.22 | 0.79 | **0.91** | **0.84** | **1.15** | **0.80** |

Looking at the machine learning models, including Random Forest Regressor (RFR), AdaBoost Regressor (AR), and Extra Trees Regressor (ETR), we observe that ETR showed the best performance with MSE = 0.62, and $R^2 = 0.89$ when using Mt and Tw features. Conversely, AR showed the worst performance with MSE = 19.80, and $R^2 = 0.28$ using Te and Ac. This shows that, in addition to the model, a good choice of features is also crucial.

Using deep learning (CNN-FCN, CNN-GRU, CNN-RNN) and transformer (BERT) leads to lower performance results than the aforementioned machine learning models, suggesting that the latter make more effective use of the combinations of features. This, in turn, motivates an analysis further delving into the features.

***Comparison of features.*** A closer look at different combinations of features shows a general tendency for combinations of more features to perform better. Indeed, when a single feature is used (rows 1-4), all models generally underperform and none of them achieves its best performance, suggesting that a single feature type does not suffice for the prediction. Combinations of either three (rows 11-14) or all four (row 15) features are the ones leading to overall best performances for six of the models (RFR, AR, CNN-FCN, CNN-GRU, CNN-RNN, BERT), with the only exception being the ETR model where a combination of only two features (Mt, Tw, row 8) performs best. Hence, where one is uncertain about the features to use, this suggests that it is better to tend towards using more features.

Apart from the number of features used in the combinations, different feature types have a different impact on the models. To analyse this, we look at the performances of the all-features setting (row 15) and the three-feature settings (rows 11-14), to see how the leave-one-out settings impact the model performance. Comparing with row 15 (all features) as the base performance scores, we observe that row 12 (removing Mt, text metadata) and row 13 (removing Tw, tweet metadata) are impacted the most with substantial reductions in performance. This suggests that both Mt and Tw are the most important features types contributing positively to the model performance. Row 14 (removing Te, textual content) has marginal impact for the machine learning models (RFR, AR, ETR) but a much bigger impact on deep learning models (CNN-FCN, CNN-GRU, CNN-RNN), which also aruges for the use of textual content as a feature in the model. Lastly, row 11 (removing Ac, account-based features) is the case with the least impact on all models, where in the majority of the cases the impact is marginal (RFR, AR, ETR, CNN-FCN, CNN-GRU) and in other cases it is still small (CNN-RNN, BERT).

Our closer look at the features suggests, all in all, that combinations of features including more feature types are generally preferred to lead to better performance, but that among the feature types, the account-based features are the ones with the least, generally marginal, impact on the predictions. Hence, this suggests that one may rely on combinations of features excluding account-based features to attain competitive performance. And this in turn indicates that account-based features are not predictive of receiving abusive replies; it is primarily the content and its metadata that can help predict if a post will receive abusive replies, rather than who the posting user is.

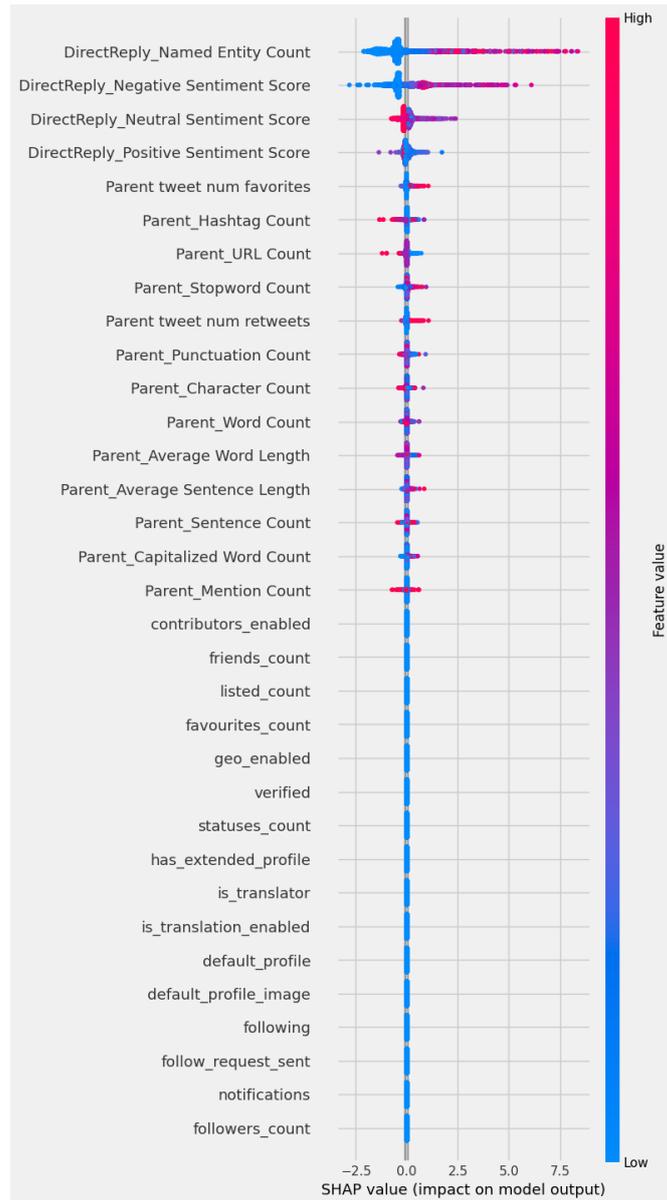

Figure 5: Beeswarm plot, ranked by mean absolute SHAP (SHapley Additive exPlanations) a summary for the Extra Tree Regressor model. The plot illustrates the impact of each feature on the prediction, with red representing higher feature values and blue representing lower values. Features are ordered by their overall importance in the model's predictions.

***Delving into the impact of features.*** Figure 5, shows the beeswarm SHAP plot, where we can identify how the underlying values of each feature relate to the model's predictions. We can see that higher values of direct reply named entity count as one of the tweet based features has the highest positive impact on the prediction, followed by the different sentiment scores of the direct replies. On the other hand, the account based features have the lowest impact on the predictions.

***Consistency of predictions.*** As we run our models on a cross-validation setting, we next look at the consistency of the performance across folds. Figure 6 shows the heatmap of the CNN-GRU model performance of the five folds including the mean. While there is some variability between folds, the average scores suggest that the model performs reasonably well across all metrics, with moderate errors and a good ability to explain variance in the data ($R^2 = 0.8$).

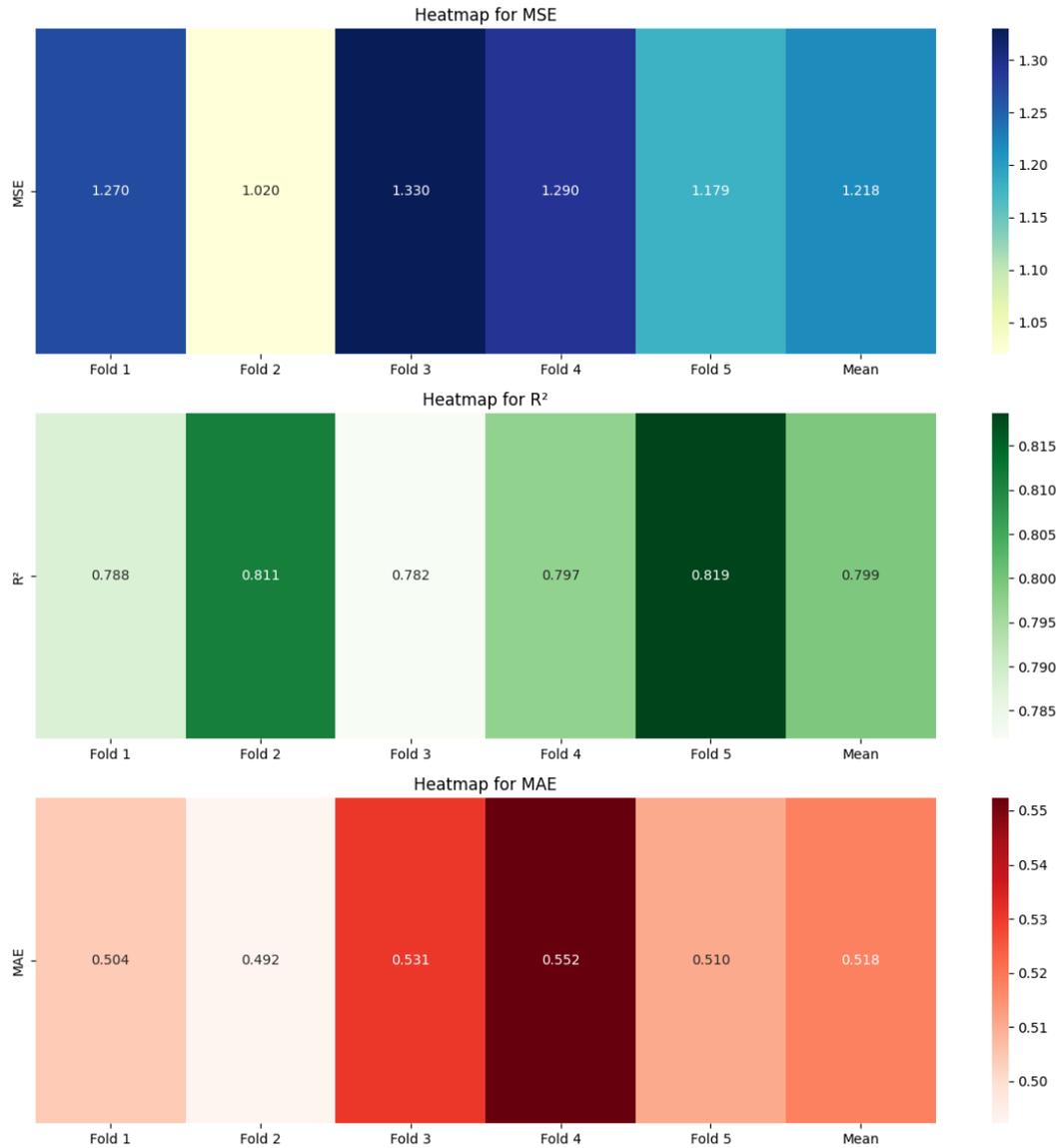

Figure 6: Heatmaps for MSE, $R^2$ and MAE - 5-Fold Cross-Validation Results. MSE is visualized with a yellow-green-blue gradient, $R^2$ is visualized with shades of green, and MAE is visualized with shades of red (dark red indicating higher values).

## 6. Discussion

In this section, we revisit the research questions of our study, further look at the features and their correlation with abusive replies, and discuss its limitations.

### 6.1. Answering the Research Questions

Here we discuss the findings and implications of our regression experiments, discussing the practical outcomes of our findings, by revisiting and answering the research questions set forth in the introduction:

- **RQ1: What categories of features available at the time of posting are helpful in predicting the volume of abusive replies a tweet will receive?**

  A set of features are available at the time of posting (pre-post), which are considered useful in providing highly accurate predictions depending on the used model. In this case, when we are aiming to provide predictions about the volume of abusive replies the user might receive,

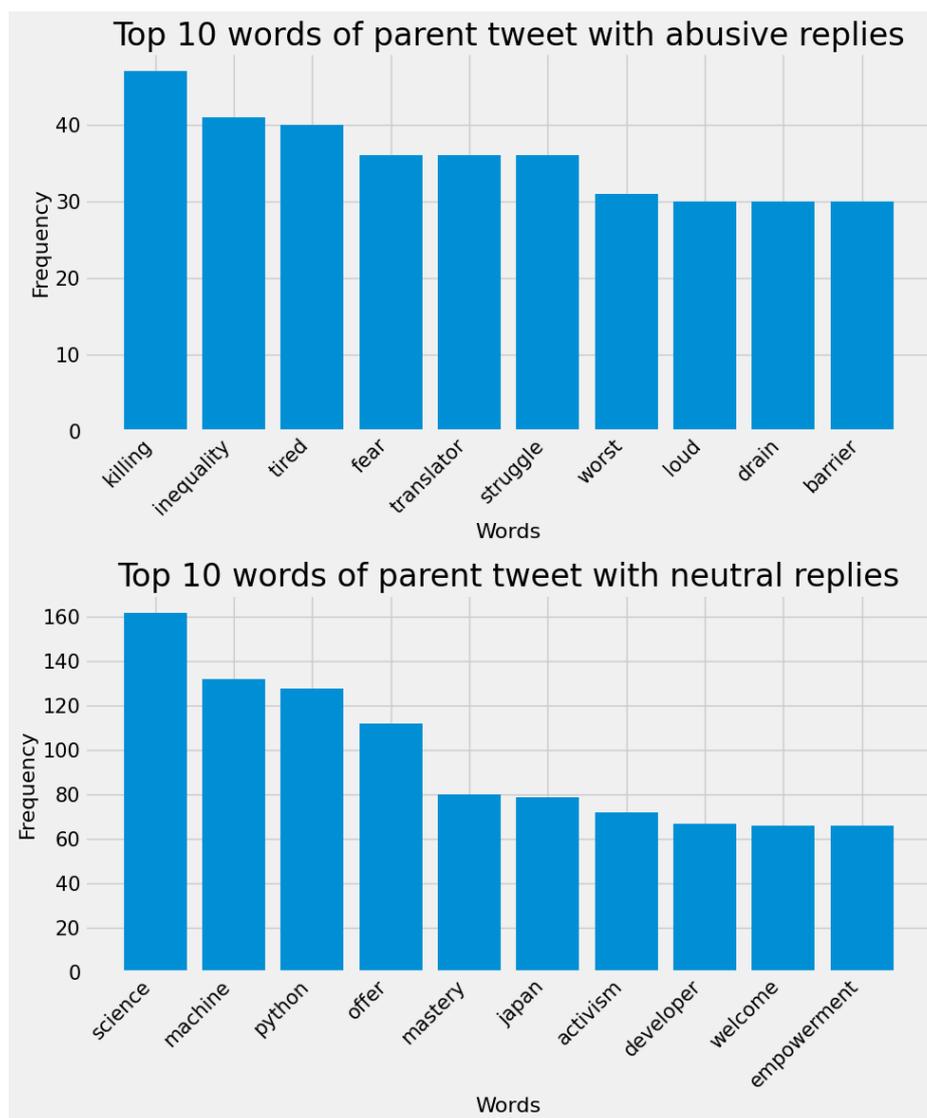

Figure 7: Top 10 words of parent with abusive and neutral replies.

The highest performance achieved is indicated by Extra Trees Regressor showed the best Performance with MSE = 0.62, and $R^2 = 0.89$ when using Mt and Tw features. While in deep learning models, the best performance with CNN-RNN MSE = 0.91, and $R^2 = 0.84$ showed when combining all features together.

Thus, based on the results of the second experiment, integrating features derived from Mt and Tw with other features leads to enhanced model performance. However, the combination of the Ac feature with text-based elements produces the lowest performance, as evidenced by the lowest R-squared value and the highest Test Mean Squared Error (MSE) among all models evaluated. This suggests that relying on the Ac-text combination is not an effective approach for improving model accuracy. The content-based features are available at the time of posting and provide highly accurate predictions on the volume of abusive replies a tweet will receive.

- **RQ2: To what extent is the prediction of the abusive replies volume linked to the identity of the user posting a message or the content of the message being posted?**

  To answer this question, we will mention how different these features are and how they are contributing to the model's performance. As the account-based features are associated with the user identity, and the other features are related to the content being posted, including the text, meta-text, and tweet-based features. In classification and regression tasks, the account-related features do not significantly enhance the models' performance. The use of the Account feature does not significantly improve the models' performance.

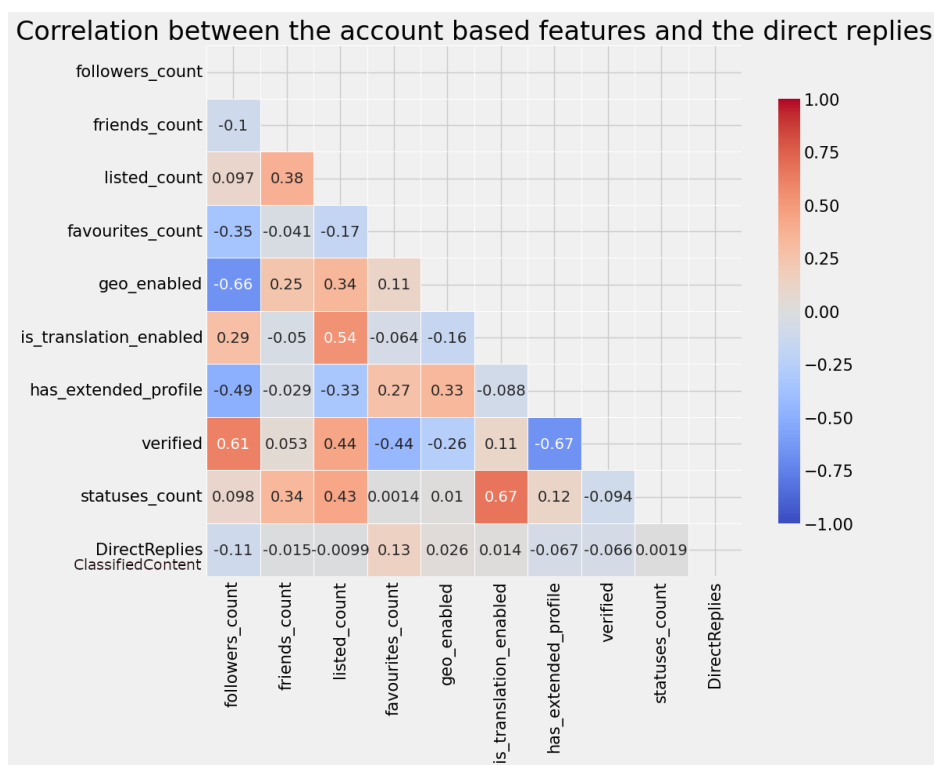

Figure 8: Correlation between the account related features and the direct replies classified content.

In the regression task, using the Account (Ac) feature alone results in the worst performance across all models, indicating that relying solely on this feature is ineffective.

### 6.2. Analyzing the Correlation between Features and Abusive Replies

Figure 7 presents the top ten words used in the parent tweets that received abusive replies as well as the top ten words used in the parent tweets that received neutral replies. This also reflects the findings that have been discussed earlier.

Moreover the features analysis provide more informative insight about the correlation between the account based features and the abusive direct replies. Figure 8 indicates very weak to negligible linear relationships between the account based features and the abusive content of the direct replies.

In Figure 9 the negative sentiment score of the parent tweet text shows a strong correlation to the abusive direct replies, and a moderate correlation between the sentence count and the of the parent tweet and the abusive replies. The neutral sentiment score of the parent tweet text shows a strong correlation to the neutral direct replies.

Thus, this stresses the fact that there are several other features but not the identity of the user who will post a message that impacts the prediction performance in both tasks. Instead, the content of the posts carries the most effective weight and proves that it can be used alone to perform accurate predictions.

Figure 5 provides more detailed explanation of the features' importance, which highlight the importance of the tweet based features with the highest ranking followed by the metatext related features, and finally the account based features with the lowest impact on the predictions in both tasks.

### 6.3. Limitations

While presenting the first such kind of study aiming to predict if a future post will receive abusive replies, our work is not without limitations, which in turn opens up avenues for future work. Our dataset contains samples in the English language only, and a study that looks at additional languages would be needed to further study generalisability in other languages (Jiang and Zubiaga, 2024).

We started by using Glove with the deep learning models as it improves it's ability to learn patterns effectively by captures word meanings and relationships. Deep learning models tend perform better with dense vector representations

Predicting the Volume of Abusive Replies before Posting in Social Media

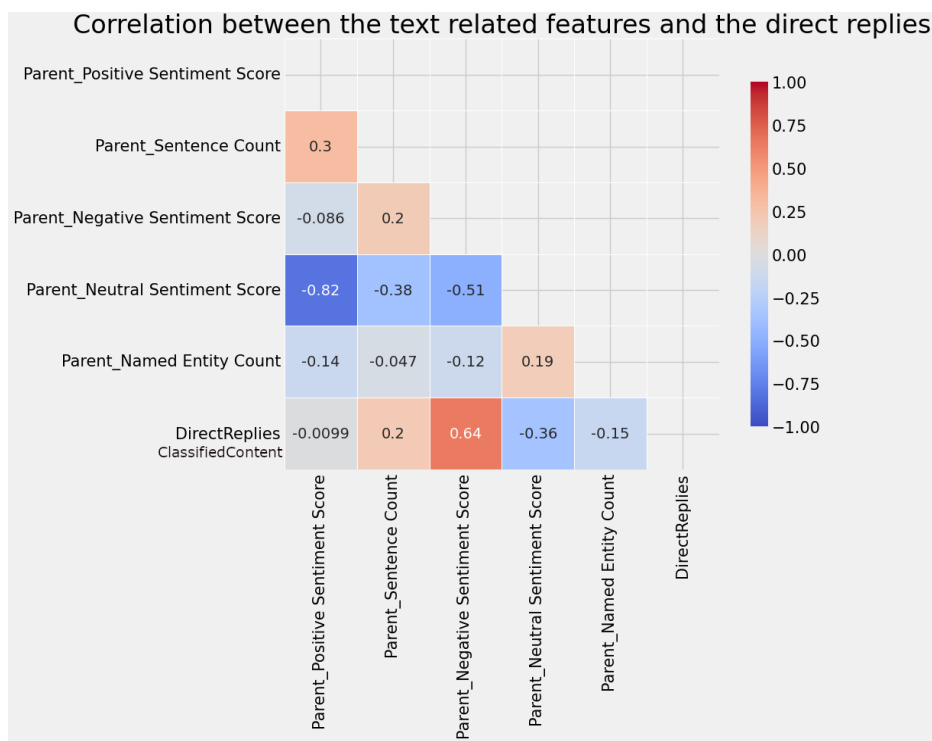

Figure 9: Correlation between the tweet text related features and the direct replies classified content.

like GloVe compared to sparse representations like BoW which we used with the Ml models to provide a form of performance comparison.

Our current experiment design accomplished the first study looking at the inherent characteristics of the content being posted and the user posting it, i.e. the target. To consider a holistic view from the user's perspective as to what might trigger abusive replies, one may need to go beyond the features studied in this work to incorporate, for example, the network of users surrounding the target, i.e. who follows them and how likely are they to post hate. Likewise, the history of the user, whether they have a prominent role in society (e.g. politicians) can play a role in determining the probability of receiving abuse. As the next step, however, one could look at identifying abusive repliers from the target's network, leading to a more comprehensive experimentation These additional features and context, while useful, are beyond the scope of this study. Instead, we perform a large-scale analysis of the features we could access.

## 7. Conclusion

Our study provides a first-of-its-kind investigation into the ability to predict the volume of abusive replies that a social media post will receive before it is posted. Using a dataset of conversational exchanges which are initiated by a post and responded to by others, our experiments investigate the ability of predicting how many of those replying posts will be abusive. This can help a user be prepared before posting in social media, and results from our experiments suggest that this prediction can be achieved with a competitive performance.

Our research findings also reveal the importance of incorporating content-based features rather than user-based features, which enable accurate predictions in the pre-post stage of a tweet. When predicting the volume of abusive replies a tweet in a pre-post stage will receive, tweet-based features are just as crucial as a combination of text, and meta-text features, while again, the account-related features do not add significant value. These findings could be crucial in providing advanced support for the content moderators in social media or other users, for example, in the governmental sector, in which such pre-post predictions could be an additional feature in the pre-post stage that provides more information and time to enhance the content.

**Ethical Considerations**

The aim of our research is to support individuals in preventing being the targets of abuse by providing the possibility for improved identification of characteristics that may trigger the abuse. If social media users have a better understanding of what may trigger abuse, and if there is an automated classifier or regression model that can help predict this with a good level of accuracy, this can help users protect themselves in advance and make an informed decision by preparing themselves in advance for possible abusive replies and take relevant actions like stopping notification, blocking abusive accounts, etc.

In studying what triggers abuse online, one of the consequences is identifying characteristics that are more prone to receive abuse and suggesting that content or users with those characteristics are more likely to receive abuse. In doing so, it is not our objective to trivialize the abuse towards content or users with those characteristics, but to raise awareness of this abuse happening and to protect vulnerable users ahead of posting.

Where a look at user characteristics can suggest that certain identities can be more prone to receiving abuse, we do not mean that any user should be concealing or misrepresenting their legitimate identities, or that abuse towards these identities should be underplayed. The aim of our study is to raise awareness of this happening in social media, to protect users where possible and to make social media a safer environment for users in vulnerable positions.

Predicting the Volume of Abusive Replies before Posting in Social MediaGandhi, A., Ahir, P., Adhvaryu, K., Shah, P., Lohiya, R., Cambria, E., Poria, S., Hussain, A., 2024. Hate speech detection: A comprehensive review of recent works. Expert Systems , e13562.

Gashroo, O.B., Mehrotra, M., 2024. Understanding the evolution of abusive behaviour in online social networks, in: 2024 3rd International Conference for Innovation in Technology (INOCON), IEEE. pp. 1–7.

Goel, A., Poswal, A., 2024. Mmhs: Multimodal model for hate speech intensity prediction, in: International Conference on Speech and Computer, Springer. pp. 95–108.

Hughes, D.J., Rowe, M., Batey, M., Lee, A., 2012. A tale of two sites: Twitter vs. facebook and the personality predictors of social media usage. Computers in human behavior 28, 561–569.

Jiang, A., Zubiaga, A., 2024. Cross-lingual offensive language detection: A systematic review of datasets, transfer approaches and challenges. arXiv preprint arXiv:2401.09244 .

Kumar, A., Maurya, M.K., 2024. Online public sphere and threats of disinformation, extremism and hate speech: Reflections on threat-mitigation. Journal of Communication Inquiry , 01968599241292623.

Kwak, H., Lee, C., Park, H., Moon, S., 2010. What is twitter, a social network or a news media?, in: Proceedings of the 19th international conference on World wide web, pp. 591–600.

Lozano-Blasco, R., Mira-Aladrén, M., Gil-Lamata, M., 2023. Social media influence on young people and children: Analysis on instagram, twitter and youtube. Comunicar 31, 125–137.

Lupu, Y., Sear, R., Velásquez, N., Leahy, R., Restrepo, N.J., Goldberg, B., Johnson, N.F., 2023. Offline events and online hate. PLoS one 18, e0278511.

Masud, S., Bedi, M., Khan, M.A., Akhtar, M.S., Chakraborty, T., 2022. Proactively reducing the hate intensity of online posts via hate speech normalization. Proceedings of the 28th ACM SIGKDD Conference on Knowledge Discovery and Data Mining 70, 3524–3534. doi:10.1145/3534678.3539161.

Masud, S., Dutta, S., Makkar, S., Jain, C., Goyal, V., Das, A., Chakraborty, T., 2021. Hate is the new infodemic: A topic-aware modeling of hate speech diffusion on twitter, in: 2021 IEEE 37th International Conference on Data Engineering (ICDE), IEEE. pp. 504–515.

Mathew, B., Dutt, R., Goyal, P., Mukherjee, A., 2019. Spread of Hate Speech in Online Social Media. WebSci 2019 - Proceedings of the 11th ACM Conference on Web Science , 173–182doi:10.1145/3292522.3326034.

Mishna, F., Cook, C., Gadalla, T., Daciuk, J., Solomon, S., 2010. Cyber bullying behaviors among middle and high school students. American Journal of Orthopsychiatry 80, 362–374. doi:10.1111/j.1939-0025.2010.01040.x.

Mondal, M., Silva, L.A., Benevenuto, F., 2017. A measurement study of hate speech in social media. HT 2017 - Proceedings of the 28th ACM Conference on Hypertext and Social Media , 85–94doi:10.1145/3078714.3078723.

Mosca, E., Wich, M., Groh, G., 2021. Understanding and interpreting the impact of user context in hate speech detection, in: Proceedings of the Ninth International Workshop on Natural Language Processing for Social Media, Association for Computational Linguistics, Online. pp. 91–102. URL: https://aclanthology.org/2021.socialnlp-1.8, doi:10.18653/v1/2021.socialnlp-1.8.

Nandi, A., Sarkar, K., Mallick, A., De, A., 2024. A survey of hate speech detection in indian languages. Social Network Analysis and Mining 14, 70.

Rawat, A., Kumar, S., Samant, S.S., 2024. Hate speech detection in social media: Techniques, recent trends, and future challenges. Wiley Interdisciplinary Reviews: Computational Statistics 16, e1648.

Ribeiro, M.H., Calais, P.H., Santos, Y.A., Almeida, V.A., Meira Jr, W., 2017. " like sheep among wolves": Characterizing hateful users on twitter. arXiv preprint arXiv:1801.00317 .
Alharthi et al.: *Preprint submitted to Elsevier*   Page 19 of 23

Predicting the Volume of Abusive Replies before Posting in Social Media

## A. Appendix

| Acronym | Full Form |
|---|---|
| CNN | Convolutional Neural Network |
| LSTM | Long Short-Term Memory |
| GRU | Gated Recurrent Units |
| RNN | Recurrent Neural Network |
| BERT | Bidirectional Encoder Representations from Transformers |
| LR | Logistic Regression |
| NB | Naive Bayes |
| RF | Random Forest |
| SVM | Support Vector Machine |
| TF-IDF | Term Frequency-Inverse Document Frequency |
| BOW | Bag of Words |
| RFR | RandomForestRegressor |
| DTR | DecisionTreeRegressor |
| AR | AdaBoostRegressor |
| ETR | ExtraTreesRegressor |

The following is a thorough description of the preprocessing procedures we implemented.

We started by converting a boolean value to binary, by translating these logical states into their binary equivalents: true becomes 1, and false becomes 0.

We also performed a **Normalization** technique to scale numerical data to a standard range, it's typically scales the data to a range of [0, 1], while standardization scales the data to have a mean of 0 and a standard deviation of 1. for the following features: Statuses count, Followers count, Listed count, Favourites count.

- Geo-enabled: Indicates whether tweets from this account are geo-tagged, meaning they include location information

- Verified: A boolean value indicating that Twitter has checked the identity of the user that owns this account

- Statuses count: The number of tweets posted by this account

- Is translation enabled: Typically refers to whether the user has enabled automatic translation of tweets in their language

- Has extended profile: Generally refers to whether the user has an extended profile with more detailed information beyond the basic profile

- Default profile: Typically refers to the default Twitter profile image used by the user

- Labelled Bio text: A short biographical description provided by the user in their profile

- Followers count: The number of users that are following this account

- Friends count: The number of users this account is following

- Listed count: The number of public lists that include the account

- Favourites count**: The number of tweets that have been marked as 'favourites' by this account

Below, we present the extensive list of features, grouped into categories, along with the exact number of features employed in each category, offering a transparent and detailed view of the feature engineering process.

### A.1. General text based features:

We performed a binary labelling of abuse, non-abuse for both the parent tweet text, and the direct reply text using the following features:

1. BOW
2. TF-IDF
3. Glove

### A.1.1. Parent tweet text based features:
1. Parent_Negative Sentiment Score
2. Parent_Positive Sentiment Score
3. Parent_Neutral Sentiment Score
4. ParentNamed Entity Count
5. Parent #
6. Parent JJ - Adjective
7. Parent NNP - Proper Noun, Singular
8. Parent NN - Noun, Singular

### A.1.2. Direct reply text based features:
1. DirectReply_Negative Sentiment Score
2. DirectReply_Positive Sentiment Score
3. DirectReply_Neutral Sentiment Score
4. DirectReply_Named Entity Count

## A.2. Meta-text based features:
### A.2.1. Parent tweet meta-text based features:
1. length_parent_tweet
2. Parent_Word Count
3. Parent_Character Count
4. Parent_Sentence Count
5. Parent_Average Word Length
6. Parent_Stopword Count
7. Parent_Hashtag Count
8. Parent_Mention Count
9. Parent_URL Count
10. Parent_Capitalized Word Count
11. Parent_Punctuation Count
12. Parent_Average Sentence Length

### A.2.2. Direct reply meta-text based features:
1. length_direct_reply
2. DirectReply_Word Count
3. DirectReply_Character Count
4. DirectReply_Sentence Count
5. DirectReply_Average Word Length
6. DirectReply_Stopword Count
7. DirectReply_Hashtag Count
8. DirectReply_Mention Count
9. DirectReply_URL Count
10. DirectReply_Capitalized Word Count
11. DirectReply_Punctuation Count
12. DirectReply_Average Sentence Length

## A.3. Tweet based features:
1. Parent tweet hashtags
2. Parent tweet symbols
3. Parent tweet user mentions
4. Parent tweet URLs
5. Parent quote status
6. Parent possibly sensitive
7. Parent tweet num retweets
8. Parent tweet num favorites

## A.4. Account based features:
1. friends_count
2. followers_count
3. listed_count
4. favourites_count
5. time_zone
6. geo_enabled
7. verified
8. statuses_count
9. contributors_enabled
10. is_translator

11. is_translation_enabled
12. has_extended_profile
13. default_profile
14. default_profile_image
15. following
16. follow_request_sent
17. notifications

More detailed explanation of the used features will be found in Twitter's documentation.[11]

---

[11] https://developer.x.com/en/docs/x-api/v1/data-dictionary/object-model/user